# Exploring the Role of AI-Powered Chatbots for Teens and Young Adults with ASD or Social Anxiety

Author: Dilan Mian (Undergraduate Student at Ontario Tech University)

**Abstract**

The world can be a complex and difficult place to navigate. People with High-Functioning Autistic Spectrum Disorder as well as general social ineptitude often face navigation challenges that individuals of other demographics simply do not themselves. This can become even more pronounced with people of that specific group when they are in their teenage years and early adulthood (that being the usual age range of college students). When they are at such a vulnerable age, they can be far more susceptible to the struggles of becoming comfortable and content with social interactions as well as having strong relationships (outside their immediate family). Concerning this, the rapid emergence of artificial intelligence chatbots has led to many of them being used to benefit people of different ages and demographics with easy accessibility. With this, if there is anything that people with High-Functioning ASD and social ineptitude want when it comes to guidance towards self-improvement, surely easy accessibility would be one. What are the potential benefits and limitations of using a Mindstudio AI-powered chatbot to provide mental health support for teens and young adults with the aforementioned conditions? What could be done with a tool like this to help those individuals navigate ethical dilemmas within different social environments to reduce existing social tensions? This paper addresses these queries and offers insights to inform future discussions on the subject.

**Introduction**

To dive deeper into this subject, there are crucial facts and points to be made first. Let us investigate what it means to have a high-functioning autism spectrum disorder and social ineptitude. An understanding of this will substantially affect the validity of the arguments in the following sections.

In the realm of medical academia, the concept of "High-Functioning Autism" is by no means an official term used for diagnosis. It is a colloquial phrase that various people use in everyday conversations. It refers to individuals who have been diagnosed with autism; however, there is a key distinction. They are distinguished from others in this condition by being able to talk, read, write, and perform essential life skills such as putting on clothes and having meals independently (WebMD, 2023).

Autistic Spectrum Disorder (ASD) is a developmental disorder of the brain with varying signs, symptoms, and effects. People who consistently exhibit the signs of this disorder are then put into the "autistic spectrum" (WebMD, 2023). Some of these people are significantly disabled while others are milder. The Intelligence Quota test scores of this vast group of people can vary widely.

For several years, individuals with severe symptoms alone have been diagnosed as autistic. Only approximately 30 years ago in the 1990s were milder symptoms of this disorder understood by people with them being known as having "Asperger's Syndrome" in the medical field (WebMD, 2023). The term itself first emerged in the 1940s after research by psychologist Hans Asperger; however, in 1994, it became an official term for diagnoses.

The use of this term would change in a colloquial sense however with the phrase "High-Functioning Autism" in recent years. Some individuals who are not medically educated may still use the phrase Asperger's. This may be due to a lack of understanding of the spectrum or an older diagnosis of Asperger's syndrome before what took place in 2013. In that year, the American Psychiatric Association (APA) placed all diagnosed individuals with symptoms related to autism into a single group, the aforementioned ASD (WebMD, 2023).

In particular, people with High-Functioning Autism are known to face challenges with communication and social interaction. Social cues/etiquette and the ability to form friendships do not naturally register with them. Social scenarios can quickly become overwhelmingly angsty, leading to a shutdown (WebMD, 2023). Eye contact with people during conversations and small talk can also be a struggle for them as well.

Some of these people can maintain their jobs, while others may not. This is largely contingent on their experience and situation. For job interviews in particular, they could very well have all the education and skills they need, but their lack of experience with long, healthy conversations could lead to an uphill battle in getting a desired position (WebMD, 2023).

It is plausible that those reading this paper have heard of or even experienced social anxiety/ineptitude themselves. Even though over 60% of people have felt social anxiety at some point (Science of People, 2023), it would be beneficial to make sure that there is no ambiguity or undermining as to what this condition entails; therefore, the means and ways it occurs will be clarified in the following analysis.

Individuals with social ineptitude face significant difficulties in socialization, carrying conversations, and interacting with people comfortably. Similar to people with ASD, this demographic commonly misunderstands social cues and feels tense in public settings. The idea of making new friends/colleagues or being humiliated in a social environment can be a source of dread (Autism Parenting Magazine, 2024).

Other notable signs of someone struggling with social anxiety are people they converse with not understanding the jokes they have uttered or even being offended by them. These same individuals usually have issues with preventing awkward silences in casual conversations; there is a fear that others may intentionally avoid conversing with them, and they overthink or regret what they say in conversations (Science of People, 2023). A few of them are comfortable in crowds but as much as individual interactions.

These are the general signs and effects of ASD and social anxiety on people at large; however, what about teenagers and young adults? Generally, they have the same symptoms as most adults, but there are unique characteristics that this specific demographic may have that others with these conditions may not have. Teens are particularly vulnerable to various stresses and socio-environmental tribulations owing to the effects of puberty (Buchanan, Romer, Wray-Lake, and Butler-Barnes, 2023). They have a much higher chance of experiencing anxiety, panic disorders, and depression. This applies to individuals who find themselves in scenarios that engender their deepest fears.

When we look at the common situations that young adults experience in college and early professional life, we can conclude that they are very prone to depression, anxiety, and panic (Asif, Mudassar, Shahzad, Raouf, and Pervaiz, 2020). This is often due to the societal and financial pressures that come with obtaining a desired college education and job in a competitive market.

This only considers the experiences of the general public in these age groups. If we factor in the added layer of people with ASD and social/anxiety, the situation would plausibly be far more concerning.

Now that we have come to understand what ASD and social ineptitude entail, as well as the common psychological issues of teenagers and young adults, what are the implications, both beneficial and harmful of using an AI chatbot via Mindstudio.ai designed to provide ethical social advice and mental support to specific people with these conditions?

In the following sections, it should be noted that young adults and teenagers with ASD and/or social ineptitude are referred to as the "targeted group" or "group in question" for ease.

**Related Works**

The name of the application is powered by Mindstudio.ai and has the title "A-Way-Out.ai". In terms of works this could be compared to, the most prominent would be the following:

ChatGPT –

To begin, below are points of ChatGPT's strengths as an application:

- ChatGPT is a very popular online general-purpose chatbot.
- It has a general framework that can understand and respond to different queries without specialized content.
- This application has a diverse conversational style to handle different subjects.
- It naturally communicates in a casual and informal tone of speech for general queries.
- It can adapt responses depending on user input.

Now below are some of the drawbacks of ChatGPT in this regard:

- ChatGPT can dispense functional responses, but it lacks deep specialization in any one specific area, including nuanced social and ethical advice.
- It does not have a substantially deep contextual understanding of specific conditions that a professional psychologist or social worker likely would.

- There is no upfront confirmation of professional psychologists or social workers having worked on and reviewed the information conveyed by ChatGPT regarding mental health support and social skills advice to the targeted group.
- There are image generation/visual aids but there is a paywall of 20 USD per month.
- While both the PC and mobile app versions of ChatGPT have text-to-speech for free for the convenience of the users, it has limitations. Both versions can replay the text-to-speech reading of prompts, but they cannot pause it and rewind or fast-forward it at their convenience. Users also cannot download the audio file to keep if they wish. This could be useful if the company powering the app was to shut down operations or if the app itself was to be removed from public use or even if the user lives in a location with an unstable internet connection.
- ChatGPT does not provide upfront reminders that the group in question should prioritize real-life counseling rather than relying on a chatbot.

Autism Support – Social Skills –

This is a created GPT which is available on ChatGPT. It seems to have been constructed by a company known as "Levanshealth". Below are points of its positive capabilities:

- It's tailored to provide support individuals with autism, focusing on social skills.
- It may have common social scenarios relevant to people with autism.
- It can issue structured scenario-based guidance to teach social skills.
- It emphasizes being supportive and educational, using a gentle and understanding tone.
- Simplified language is utilized to ensure clarity and ease of understanding for people with ASD.

Now below are some noteworthy criticisms.

- This GPT does not necessarily do much to distinguish itself from just utilizing ChatGPT. To be more specific, it uses Dall-E to generate images that can help understand socially beneficial things like proper posture and more. There does not seem to be anything stopping regular ChatGPT from doing the same thing if someone were to ask for it.
- There is no clarity as to whether the chatbot was approved of or at least recommended by professional psychologists or social workers during the chatbot's development phase. There is no easily accessible information on who did recommend it such as a LinkedIn profile containing their contact information and qualifications.
- Similar to a general-purpose chatbot, it may not be deeply personalized or cover a vast range of ethical dilemmas and mental health support comprehensively in a way that a real social worker or psychologist would.
- It does not clearly state that the user should prioritize seeking in-person counseling rather than only relying on the application.

Wellness Buddy –

This is an AI application targeting Kenyan university students with a focus on mental health support, especially for anxiety and depression, in low-resource settings (Ogamba, Gitonga,

Muriithi, Olukuru, & Sevilla, 2023). It has a primary goal to address the stigma around mental health in Kenya. Below are some positive functionalities for this app:

- It has a functioning chatbot that is tailored to anxiety and depression, with added integration into the Kenyan cultural context.
- The app has a focus on usability and engagement with users.
- It has a strong reliance on transfer learning via neural networks.

Now below are some potential shortcomings of the application:

- It has a somewhat limited scope of mental health concerns addressed.
- The app has a potential overfitting of models.

MIT Study-Based Application –

This is an application focused on mental health support and improving social interactions through AI chatbots, specifically with empathetic engagement (Shen, DiPaola, Ali, Sap, Park, & Breazeal, 2024). Beneath are some of its positive traits:

- It uses generative AI with leveraged models like GPT for user interactions and addressing certain needs.
- It highlights concerns around transparency, accuracy, and the risk of the overuse of AI for mental health support.
- The study includes methods for assessing effectiveness with user feedback and professional evaluation, with an emphasis on ethical considerations in design.

Below are some differences between this app and my proposed Mindstudio application:

- The Mindstudio app will be tailored for specifically adolescents and college-age people with high-functioning ASD or social anxiety, focusing on their unique challenges in socialization and ethical decision-making. The MIT application in contrast broadly explores empathy in AI interactions without targeting a specific demographic and their unique needs.
- In terms of customization and features, the Mindstudio app will have personalized role-playing scenarios and accessibility features such as downloadable text-to-speech or a menu-driven interface tailored to user needs. In contrast, the MIT application is focused on controlled crowd-sourced studies to measure empathy levels and examine qualitative feedback.
- Regarding technological integration, "A-Way-Out.ai" will be built with a practical implementation to focus on tools like menu nodes, workflows, and visual aids. Meanwhile, the MIT app leverages fine-tuned AI models for story retrieval and generation to evaluate user reactions.

Saarthi –

This is an AI chatbot focused on delivering Cognitive Behavioral Therapy (CBT) and remote health monitoring for people with anxiety and depression. It is a general mental health tool targeted at a broad audience seeking accessible, evidence-based treatments (Rani, Vishnoi, & Mishra, 2023). Below are some of its strengths:

- Saarthi has therapeutic techniques and information through Natural Language Processing.
- It integrates CBT to address mental health concerns.
- There is a community platform included to connect with peers and professionals.
- Natural Language Processing is used for personalized support.

Now below are some of the differences between Saarthi and "A-Way-Out.ai":

- Regarding Methodology, Saarthi is focused on applying machine learning techniques like N-Gram, TF-IDF, and sentiment analysis for text generation. It also relies on Rasa for advanced conversational abilities. "A-Way-Out.ai" in contrast will use a node-based interface from Mindstudio for non-programming-based functionality. It will have GPT-based responses with downloadable text-to-speech and image generation. The effectiveness of the Mindstudio app will be evaluated with professional social worker or counselor feedback.
- Saarthi aims to improve CBT, addressing stigma and financial limits. The application acknowledges that AI cannot replace human counseling, particularly for mental health support. Meanwhile, the Mindstudio app will highlight the importance of real-world counseling and clarify this in the app interface. It also seeks to prevent excessive reliance by promoting limits on usage and encouraging real-life interactions.
- While Saarthi has limited personalization for niche demographics like the group in question and its general focus may dilute its effectiveness for certain mental health issues, "A-Way-Out.ai" is wholly targeted towards the targeted group.

Mental Health Support Chatbot (Gupta et al.) –

This AI application is designed for generalized mental health support addressing anxiety, depression, and stress with Natural Language Processing. It is generally aimed at people worldwide without emphasis on specific and niche groups (Gupta, Joshi, Jain, & Garg, 2023). Below are some of the strengths and positives of the project:

- This app uses deep learning and natural language processing to process user inputs.
- It has pre-filled responses and tracks user mood.
- TensorFlow and Keras frameworks are utilized for software architecture.
- The implementation of the project uses diary logging, sentiment analysis, and habit tracking.

Below are some key differences between the general Mental Health Support Chatbot mentioned above and "A-Way-Out.ai":

- Gupta's chatbot is constructed from scratch with Python coding language that emphasizes natural language processing. The Mindstudio app on the other hand will use a no-code platform, using pre-designed workflows and nodes for quicker, easier-to-understand development.

- While the Gupta chatbot emphasizes conversational empathy and emotional support across broad mental health categories, the Mindstudio app is focused on certain tools for social skills, ethical advice, and roleplay scenarios for people with ASD or social anxiety of a particular age range.
- The Gupta application garners feedback through NLP performance metrics rather than professional input/feedback from a social worker, counselor, or psychologist, which is the evaluation methodology of the Mindstudio app.
- Gupta's chatbot briefly addresses ethical concerns but there is a notable lack of professional oversight, raising questions about how a real counselor, social worker, or psychologist would feel about the app's capabilities for its intended purpose.

My Mental Pocket –

This is a chatbot project aimed largely at university students and faculty with an overall focus on handling depression and managing emotional health (Park, Kim, 2023). Below are some of the key differences between it and the Mindstudio chatbot:

- While My Mental Pocket is focused on university students and staff broadly for mental health support, the Mindstudio chatbot is focused on adolescents and young adults with High-Functioning ASD and/or social anxiety. The Mindstudio project addresses mental health concerns and social skills advice, with a focus on ethical decision-making and interpersonal interactions for that particular group of individuals.
- Features such as "Pocky", an AI chatbot, and wellness games are used to diagnose depression indicators like cognitive control with My Mental Pocket. On the other hand, "A-Way-Out.ai" will use tools such as conversation roleplays, visual aids, and ethical decision-making frameworks tailored to ASD-related social struggles. Accessibility will be emphasized with the Mindstudio app, with user-friendly features like image generation. Another key one is downloadable text-to-speech files that have convenient fast-forward and rewind tools.
- While the scope of My Mental Pocket is broad, it potentially may dilute the depth of guidance for specific user demographics like ASD individuals. It also has heavy reliance on theoretical models which may limit practical insights for real-world challenges. It also lacks the professional input or review of people like counselors or social workers.

Jennifer Chatbot –

The Jennifer chatbot was designed as an information portal during the COVID-19 pandemic to provide credible, easily accessible, and timely updates about the virus and public health measures (Xiao, Liao, Zhou, Grandison, & Li, 2023). Below are some of its attributes and capabilities:

- It was built with an expert-sourcing framework with over 150 scientists and health professionals.
- It has extensive manual QA curation, with regular updates based on expert reviews and changing scientific knowledge.

- The app provides fact-checked, evidence-based responses to user queries with automated updates for certain categories like statistics.
- It emphasizes interaction focused on reducing misinformation and gaining trust through transparency.

Below are some of its drawbacks and differences from "A-Way-Out.ai":

- It is resource-intensive due to reliance on experts for content creation and validation.
- Updates for public policy and evolving disease knowledge depend on manual efforts.
- Its focus is solely on COVID-19-related information, with less adaptability for broad mental health concerns or user ethical concerns.
- Jennifer can give general answers but lacks social skills support or customized roleplay scenarios.

**Proposed Solution**

There is a plethora of ways that this potential application could distinguish itself positively from the other applications and studies mentioned above:

- This proposed application will have the professional input of at least one social worker who specializes in mental health support for the targeted group and coaches them on social etiquette during the app's development process.
- The approved responses used as input will address complex social and ethical dilemmas from the group in question while being reviewed by at least one qualified social worker, counselor, or psychologist. The different tools the app provides will be laid out in an organized, interactable menu.
- It will be catered specifically to the common interests and experiences of teenagers and young adults (late teens to early 20s). The chatbot will be tailored to their particular needs and challenges.
- This chatbot will include accessibility options such as text responses and visual aids. Audio-based interaction can be very calming compared to simple text readings for stressed users, especially if the AI allows for a realistic speaking voice. Visual aids are also an easy way to engage users who may not necessarily want to just read. Imagine someone from the group in question looking for visual examples of good posture when having conversations. It is important to note that ChatGPT presents a paywall of USD 20 per month to use image generation capabilities. This also applies to all the GPTs independently constructed there as well. Individuals of the targeted group (especially those in high school from impoverished environments) may not be able to afford such services. The text-to-speech responses can be paused, rewound, fast-forwarded, and downloaded at the convenience of the user. Downloading them in particular could be very useful if the application were to be removed or cease to operate for any given reason. If the user lives in a location with unpredictable internet service, simply being

- able to quickly download audio responses with valuable counseling-related information presents the benefit of accommodation.
- Personalized roleplaying scenarios for ethical decision-making will be prioritized with clear choices for female or male roleplays depending on the user's preferences. The chatbot will give the option for roleplaying as a classmate, teacher, co-worker, or workplace boss.
- The application will clarify upfront in each provided tool that is accessible via the main menu that the app is not meant to replace real social workers or psychologists, rather they should be sought first as a priority.
- Mindstudio requires the user to create an account by manually entering their email address and creating a new password. If they doubt the privacy and security of Mindstudio regarding their customized password, they can use their Gmail account and respective authentication from that avenue instead. These are the methods of protecting the accounts of people using and especially those creating applications via Mindstudio.
- Data usage on Mindstudio involves the developer having a set budget in their workplace for AI applications being run by the company. This budget can be used by the users to interact with the created chatbots. Each AI-generated response uses tokens which all cost money, therefore subtracting from the workplace budget. The developer has the freedom to place limits as to how much they can use of the workplace budget before they can be granted access to the application again.
- At least one professional and qualified social worker or counselor will be required to review the application during the design phase to give tips on how to make the Mindstudio chatbot as effective as possible. The same individual(s) will be asked to review the app once more after all proposed features have been implemented to design the overall impact the project has for its intended purpose.

**Methodology of Building the App**

Mindstudio Architecture – The chatbot has been constructed in Mindstudio as stated before. It has an interactable main menu (image below) for engagement and clarity of what it can do for the user. Mindstudio utilizes nodes in its automation to perform different functions without programming. This added to the ease of the app's creation. The menu has a starting node which uses a menu tool to connect to different portal nodes (Figure 1). These portal nodes known as "Jump" are linked to different workflows that contain the user-machine interaction process for different purposes. These purposes include advice related to ethical decision-making, social skills support, mental health, resources (i.e. organizations that work with the group in question for mental health and social skills support), visual aids (for images used to demonstrate helpful advice), and conversation roleplays.

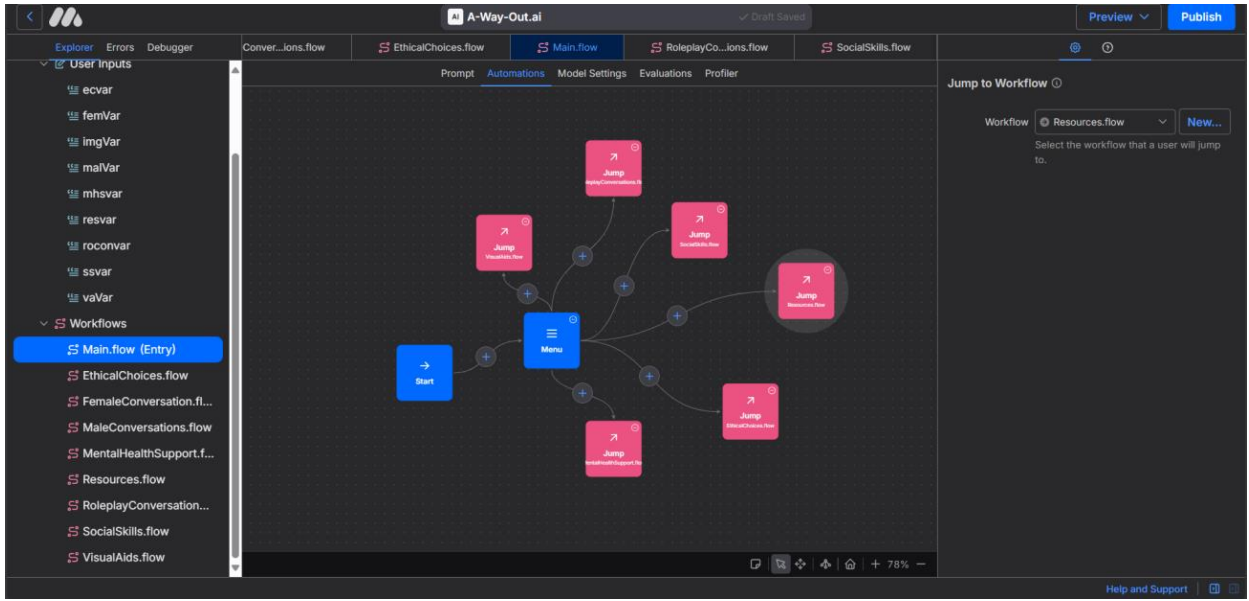

Figure 1: This shows the main menu workflow in Mindstudio.

The main menu (Figure 2) in the actual application appears with the text introducing the app's purposes and nature. It also contains an explicit reference to the social worker who reviewed the application, their qualifications, and the direct link to their LinkedIn profile so the user can reasonably verify who the individual is and the person's authenticity.

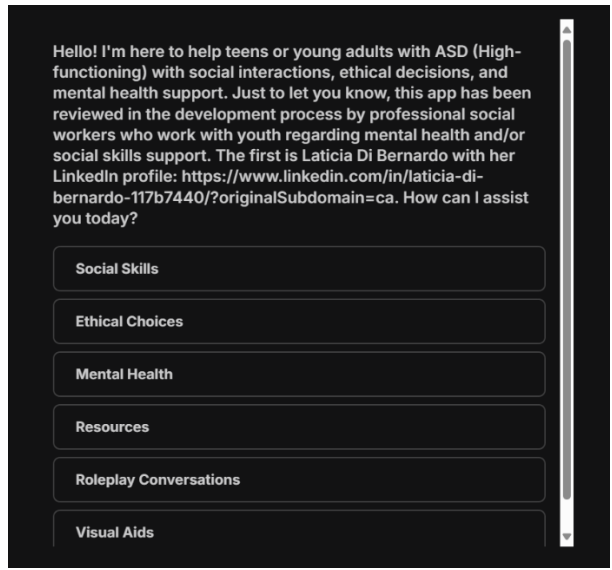

Figure 2: This shows the main menu from the user perspective when running the app.

The Ethical Choices, Social Skills, Resources, Mental Health, and Conversation Roleplay sections have a very similar workflow. The defining difference between them is that each workflow deals with respective concerns that are made clear in the opening prompts of all of them. The Ethical Choices workflow for example contains a starting node to begin it, the user is given the ability to use a variable called for this part of the app "ecvar" (Figure 3). This variable is used in a User Input node and referenced in a succeeding Generate Text node that represents the introductory text of the workflow.

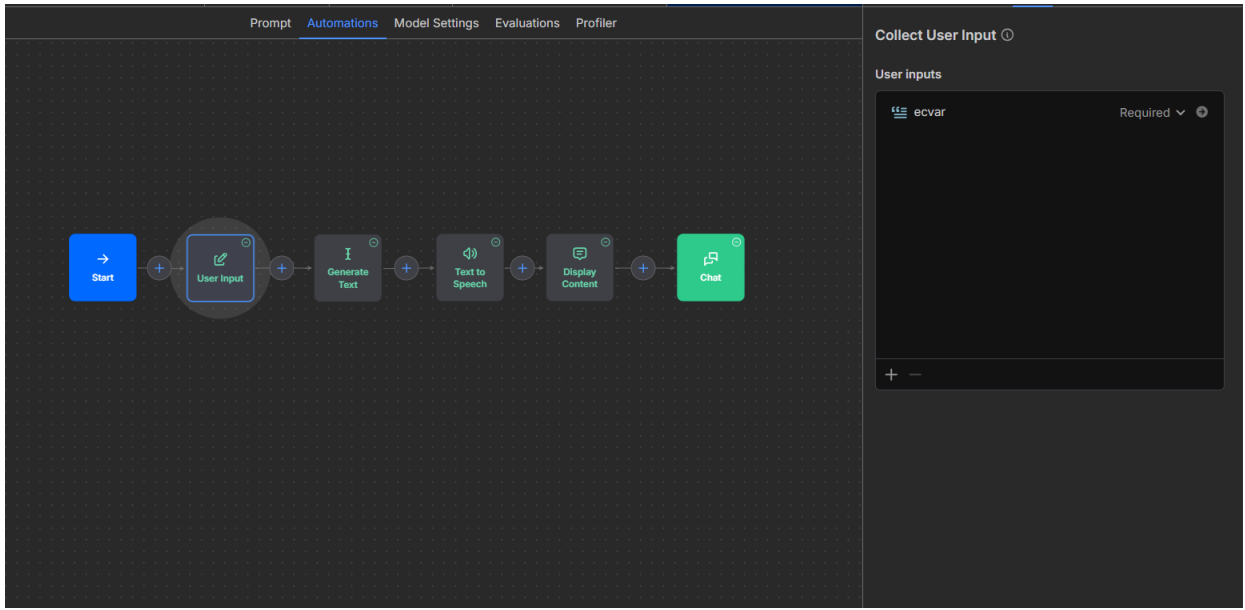

Figure 3: This shows the Ethical Choices branch workflow

The introductory text is about the purpose of this branch of the app with kind words to express its focus on discussing the ethical dilemmas of adolescents from the targeted group. The text is then followed by a reference to the "ecvar" mentioned before as this will then take the user input as the response to the app's initial prompt (Figure 4). Below that section labeled as "Prompt" is the Settings which include how the AI-generated response should be stored, in this case, it is saved as the variable "ecResponse" to be used in the succeeding Display Content node.

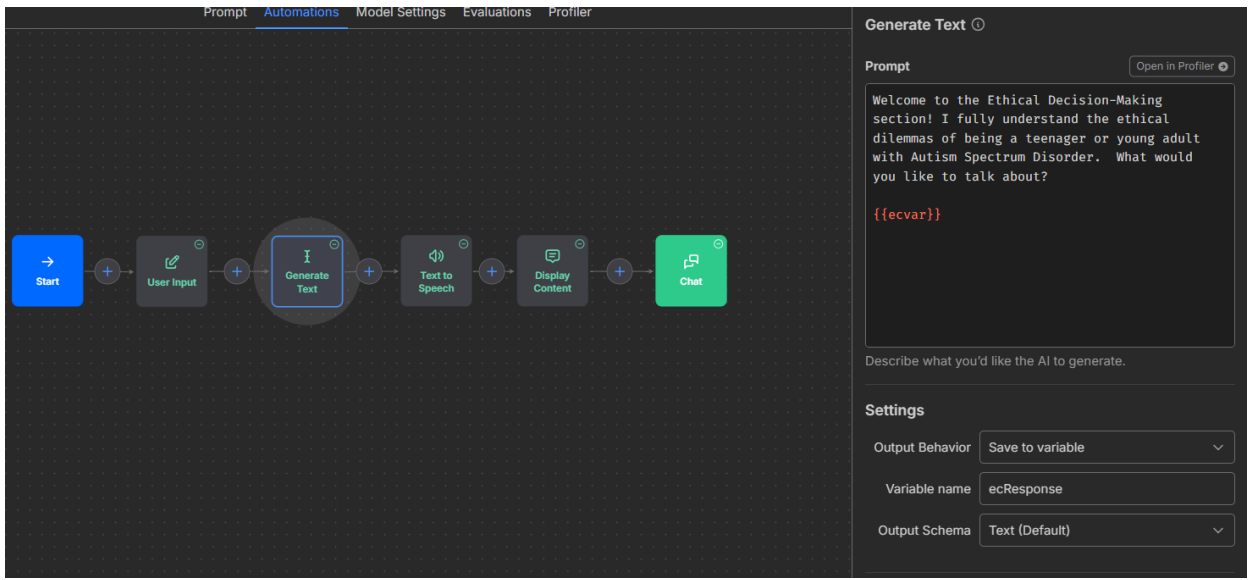

Figure 4: This shows the Ethical Decision section's Generate Text node contents.

Beneath that, there are the Model Settings (Figure 5). These include which LLM model should be used to generate the response to the user prompt (this case being GPT-4o Mini) along with the Temperature and maximum tokens allowed to be used for the response. It ends with a reference to the workflow this prompt is intended for.

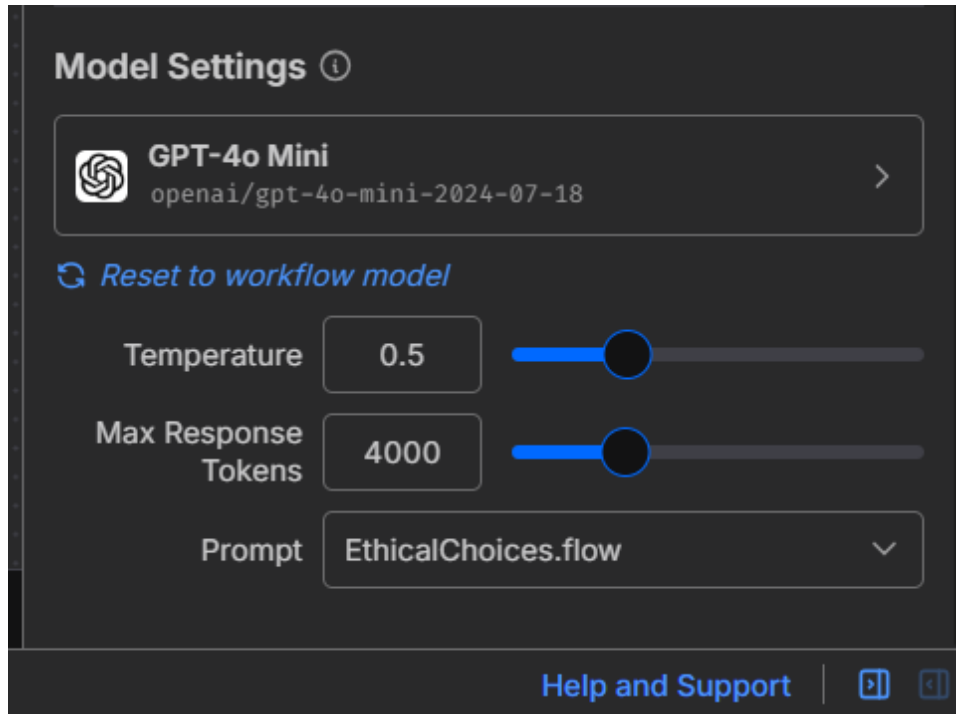

Figure 5: This shows the model settings for the chatbot's response capability.

Following that the Text to Speech node allows for the AI-generated response to have the implied text-to-speech functionality. The Mindstudio interface has the ecResponse variable stored as the text to be used by the node with its output into the Display Content node being represented by the "ethnicAudio" variable. Then beneath that there is the specific text-to-speech model being used for this functionality which in this instance is from OpenAI (Figure 6). Lastly, there is the voice option, which includes a variety of voices of different pitches. Shimmer was the one chosen here.

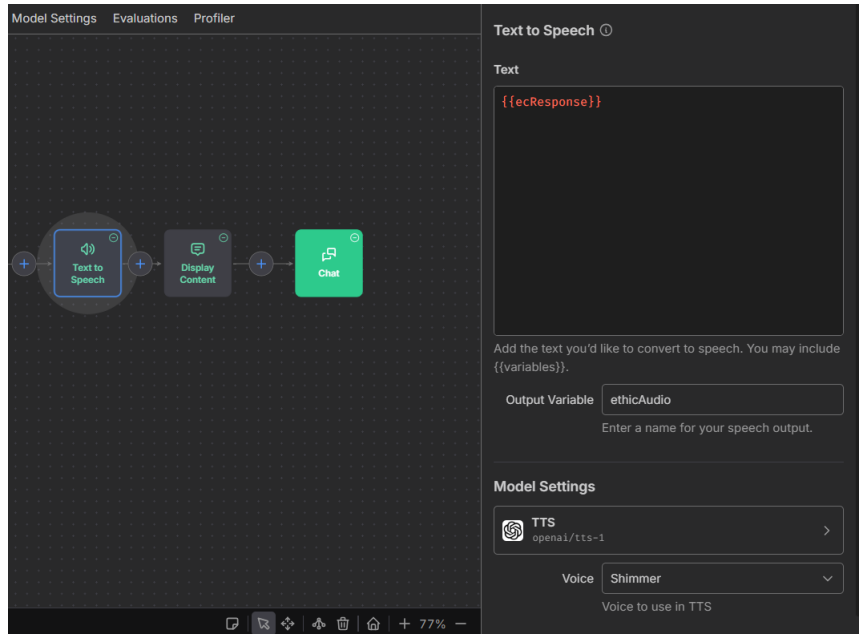

Figure 6: This shows the contents of the Text-to-Speech node in the same workflow.

The most important node after this is the Display Content one. This node facilitates the text-to-speech to be presented to the user with the AI-generated text response in full display. The node accomplishes this with HTML syntax for the establishment of the ethicAudio variable's contents mentioned earlier with the printed content of ecResponse viewable along with it (Figure 7).

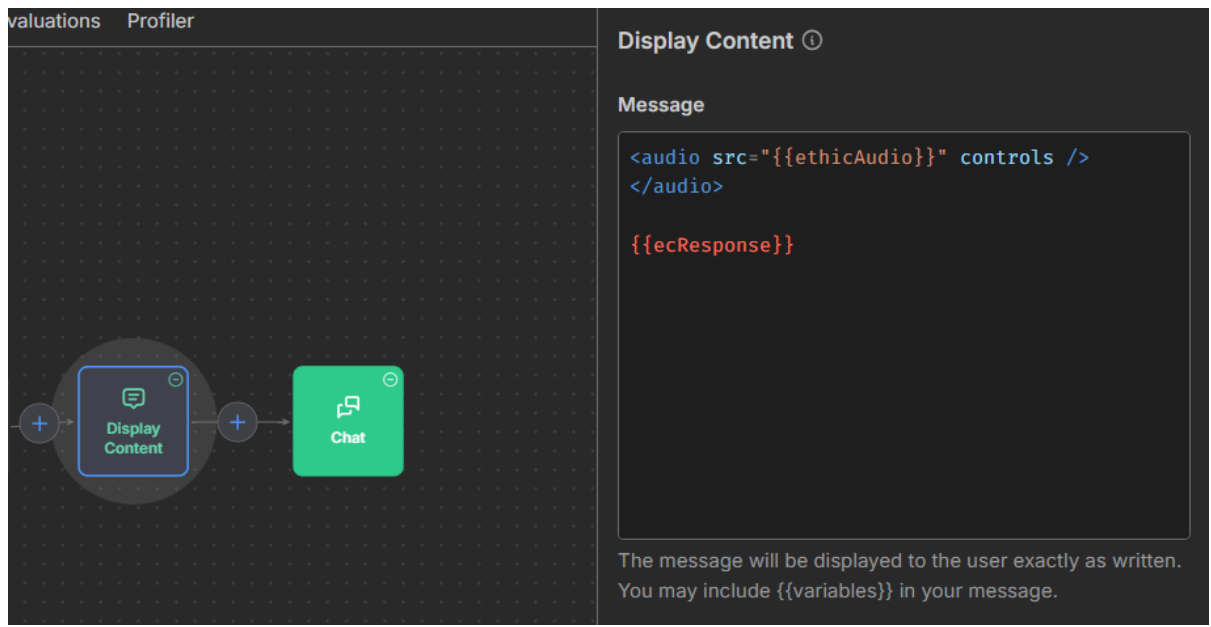

Figure 7: This shows the contents of the Display Content node in the same workflow.

The final node seen above is the Chat node which terminates the initial interaction process allowing for the user to respond with another prompt in a loop of sorts.

This is the workflow architecture for most of the sections of the application. The only exception would be the Visual Aids section. The workflow for that section utilizes a node called Generate Image rather than the Text to Speech one. Ask it can be seen in the image below; this node uses the variable representing the response of the chatbot (in this workflow it is called visualImg) to the user prompt to generate an image rather than text-to-speech. The image is stored in the variable displayImg and beneath that includes the generation model settings. This includes which model is to be utilized, which for this workflow would be OpenAI's Dall-E 3 (Figure 8). Then the aspect ratio and pixelization quality are chosen (in this instance 1024x1024 and Standard were used respectively).

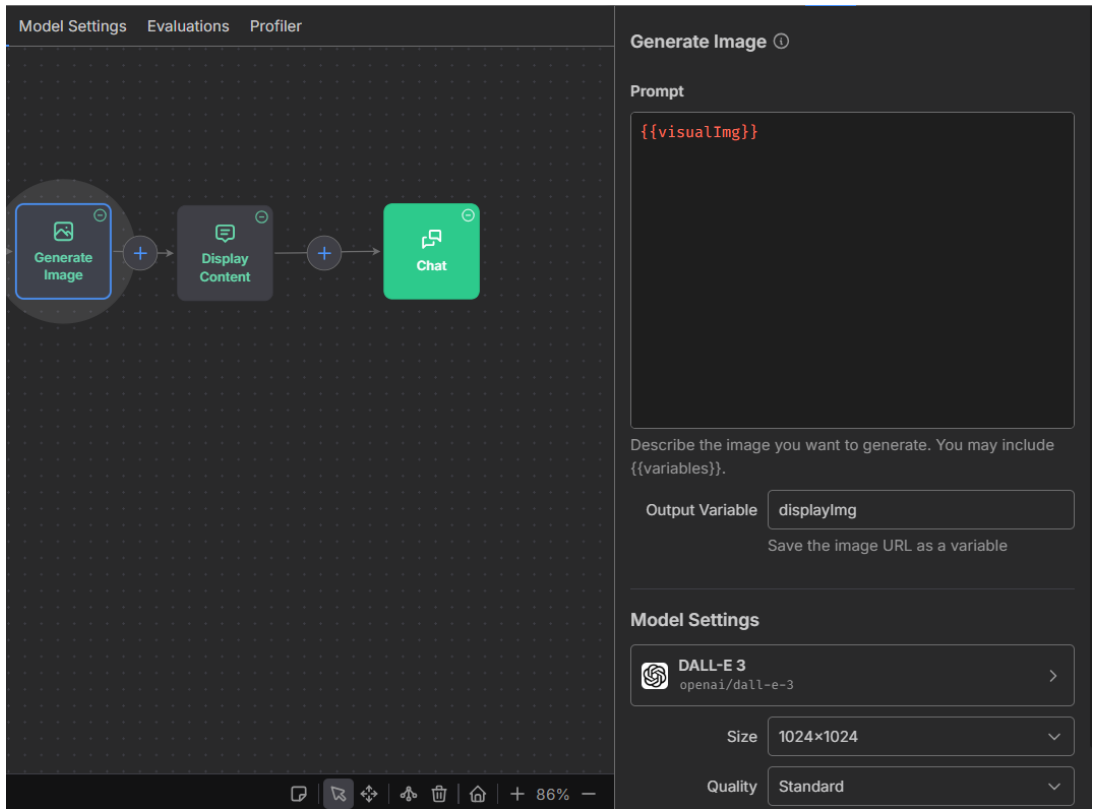

Figure 8: This shows the content in the Generate Image node in the Visual Aids workflow.

Following that is the Display Content node shows the image using a different syntax than the text-to-speech node (Figure 9). Mindstudio in this case uses an exclamation mark and various brackets and braces to convey the generated image on screen with the displayImg variable. The last section of the node's content has the visualImg displaying generated text describing the image.

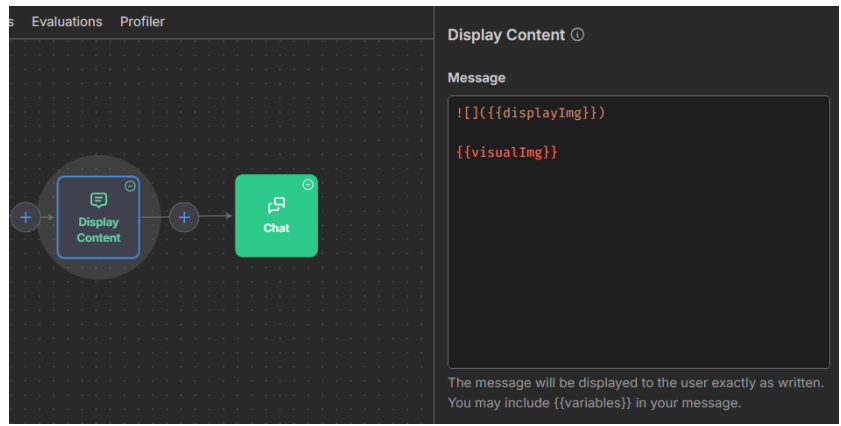

Figure 9: This shows the contents of the Display Content node in the Visual Aids workflow.

Another important section of the application is the roleplay conversations section mentioned before with both female and male character options (depending on the user's preferences). These two branches utilize female and male-sounding voices respectively. In Figure 10, the workflow for this section has a menu node which first has a prompt of what text the menu will use to introduce itself. The Menu node then has two options referencing the two respective branches to take the user to.

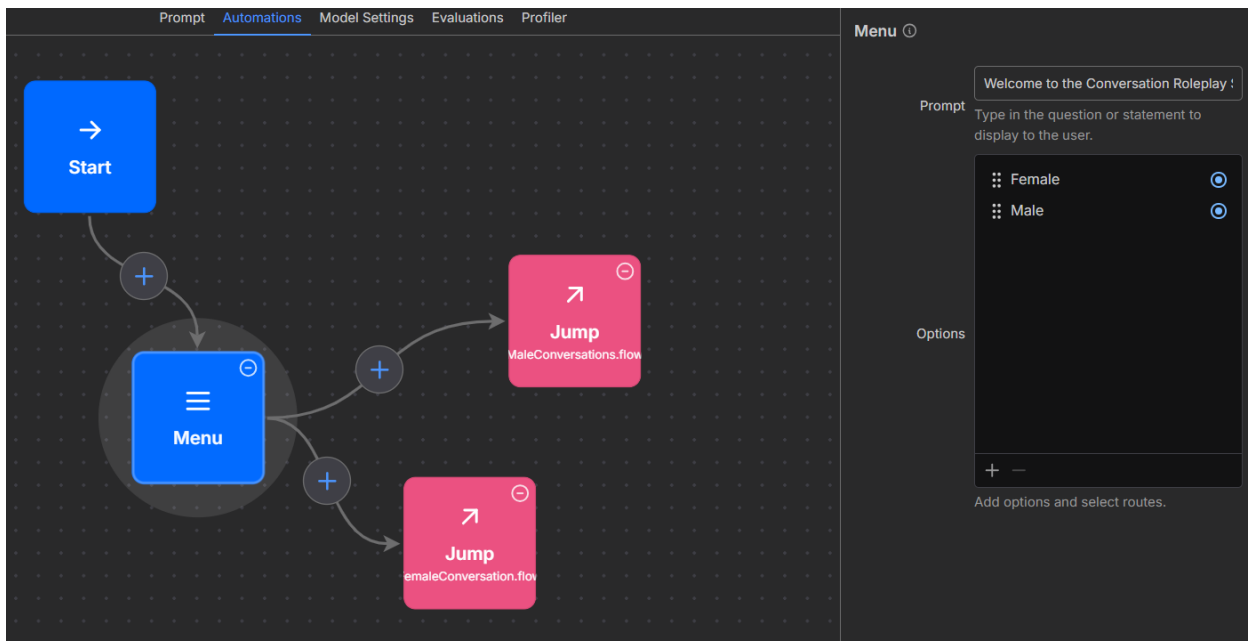

Figure 10: This shows the contents of the Menu node for the Conversation Roleplay workflow.

In both the male and female workflow branches, they have the same logic and structure as the Ethical Choices workflow except the variable names (i.e. user input and generated responses) are distinguished for the former two workflows.

Evaluation Plan – The overall effectiveness of the chatbot once it is implemented will be evaluated via interviews with at least one qualified and professional social worker or counselor on their experience using the app in its varying sections. To clarify, once this is done, the application will not be used for regular usage by organizations dedicated to helping the targeted group due to the financial backing required for the app. Mindstudio requires developers to pay a fee each month to run the application, and if no organization is willing to buy the app and run it themselves or pay the developer to keep it published, the project will have to be taken down after this paper is completed. The main purpose of this research project is to understand the ways AI can benefit the targeted group and if it can create professionally approved solutions for them (via avenues such as Mindstudio.ai).

**The Benefits**

Table 1: This is the summary of all the benefits of the application –

| Benefit | Explanation |
|---|---|
| Social Skills Development Support | Individuals of the targeted group will get intelligent tips on understanding tips on ethically-sound communication. |
| Understanding Social Cues and Communication | Individuals of the targeted group will be able to comprehend non-verbal communication from peers. |
| Ease of Access | Users of the app will have free access to it and its features regardless of their locality as long as they have somewhat working internet access. |
| Support With Addressing Trauma | The app can present temporary tips for addressing traumatic situations while not replacing actual counselors and social workers who should be considered first. |
| Open To Improvement and Innovation | The app if successful in its promises can open the door for researchers and engineers to look for new ways to introduce AI apps to younger children with ASD to provide social skills support for them before they enter adolescence. |

Social Skills Development Support – If the engineers behind an AI chatbot could conjure one that was capable of understanding the deepest and most pressing issues for the targeted group while being seen as credibly safe to use, it would, by all means, be an up-mountain battle (uphill would be an understatement). For example, if a government-funded academic institution like a university or high school were to utilize this chatbot, it is unlikely that it would even happen unless the data for the AI system were reviewed or made by professional psychiatrists, especially those who have studied the many psychological struggles of teenage life. In this way, the chatbot could be trustworthy and free of the dangers of inaccurate information (more on that in the limitations section). If this chatbot functions to have the most accurate information it can, many would undoubtedly be drawn to it as a source of information on navigating uncomfortable social interactions. Thus, whenever the targeted group utilizes a chatbot with this degree of knowledge and specialty, it is plausible they would receive the type of advice that could free them from many of the psychological trappings of everyday interactions. As a simple example, let us say that there is a 14-year-old boy in high school named Richard, who has ASD. Richard goes to the library to study for a test and he notices how his peers have a copy of the textbook they all (including Richard) needed for the test while he did not. If Richard had not been in a situation where he needed to ask to borrow something from people he was unacquainted with, it could have been a source of confusion or doubt. If only Richard had access to the chatbot on his phone, he would explain his circumstances and what he was looking for. The chatbot would swiftly give him helpful and professional-based advice on handling the situation ethically, one that is most conducive to social cohesion or bridge-building. One more example could be a college student named Ria who struggles with social anxiety as a product of her introversion. If Ria was meeting peers in her class for the first time, and was curious about their likes/dislikes, but did not know how to ask about them in a way that would be socially acceptable, this would likely lead to self-doubt or increased discomfort. If Ria asked a simple question about the hobbies of a peer, but the phrase included somewhat intrusive wording, it is possible it would not bode well. Her peer may be passively offended and cease attempts to continue conversing with her. Had Ria been given access to a chatbot on her phone or laptop that could quickly tell her how to convey questions in a socially appropriate manner and also respond in a fitting way to passive-aggressive behavior, she could have either avoided that situation altogether or found a way to repair it when it went south. The natural conclusion of all this is that an AI-powered chatbot that has been professionally reviewed by psychologists or social workers from appropriate organizations could be a very easy tool for advising on dealing with confusing or troubling social situations.

Understanding Social Cues and Communication – Building from the previous point, a chatbot with appropriate guidance for the group in question could help them understand the mindset and perspective of the people around them. If someone with ASD were about to start their first day of high school, a fitting AI system could reasonably give them rapid-fire information on understanding the social expectations and environment they would likely have to meet and interact with. The person from the targeted group can learn the most essential facts about how their peers might treat someone like them and some helpful tips to interact with them in a way that is conducive to social cohesion given all the ambiguities they have internally. Through continuous practice with real-world interactions and scenarios, they could gradually become accustomed to applying the essential lessons from the AI chatbot and then increase their knowledge of the nuances of how people converse and express themselves. From the reverse angle, if a student in a high school or college who is not part of the group in question wanted to learn more about the feelings of one of their classmates who are in the targeted group, then the AI chatbot could supply quick and professional information on that front. AI systems such as this can open the door to both unity and social bonding. There is also the positive outcome that through the targeted group using the app for this purpose, they can easily blend into the point where people outside the targeted group (especially those without ASD of any variant) will see just how human and how similar we all are. The fact that we all have our own struggles and challenges in which our inner strengths come to light. People can have different trials and tribulations, but they do not deem one group as socially or intellectually inferior to another.

Ease Of Access – It is no mystery that Artificial Intelligence has few if any, limits in terms of access to anyone with an Internet signal. Judging by the fact that the majority of the targeted group has the aptitude to use computers, search engines, and other tools that would provide access to the plethora of AI systems online, logically, they would not have a challenge in finding an AI system that can help them deal with their day-to-day dilemmas. According to professional resources, it seems very likely that with the widespread empire, which is the digital age, AI applications could have a much quicker reach than in-person psychologists and doctors (Ettman, and Galea, 2023). Imagine people amongst the group in question who happen to live in financially challenged regions of the globe such as South Asia, much of Sub-Saharan Africa, Eastern Europe, the Middle East, or even in lower-class neighborhoods in the Western Hemisphere. The reach of people in the targeted group includes those who come from families primarily concerned with maintaining basic necessities rather than having the privilege of affording expensive mental health support services. The establishment of a professionally reviewed app that could provide mental health advice as well as ethical tips for social dilemmas would be a major convenience for people from the group in question who lack the same economic privileges as the people who initially made the app in the first place. For many years, access to mental healthcare was mostly out of reach for people of the lower class and impoverished environments (Hodgkinson, Godoy, Beers, and Lewin, 2017). There is evidence that impoverishment can facilitate the fracturing of one's own mental health. When we apply that to the experiences of the targeted group, there is a lot of damage that can be done. People in the group in question from a lower-class neighborhood or region can have unique experiences compared to their neighbors, with their own mental health challenges and needs. We should keep in mind that impoverished neighborhoods can be correlated with higher crime rates (Toronto Security Company, 2023). This includes gang activities, a recipe for a hostile and stressful environment (National Gang Center, 2020). The targeted group would likely be much more prone to anxiety and alienation in a place where their safety is at risk when in or outside their house along with the unfriendly gang-like activity in their area. Adding to the fact that the person's parents are likely unable to afford high-standard therapists or social workers to regularly give them ethical advice specific to their conditions/needs, there is a lot of room for their livelihood to fall apart. If the specially designed Mindstudio app were made widely available for them for free via the internet, it would be a very convenient way to supply them with the temporary, helpful, and arguably necessary material for dealing with mental woes or social anxiety (the potential absence of privacy and confidentiality will be discussed in the harms section). There is also a threat to the people of the targeted group being conditioned by gang culture in these types of neighborhoods to where it becomes normal or even admirable. Gang culture, in general, can be very attractive to youth (National Gang Center, 2020). This is again keeping in mind that the group in question tends to lack knowledge of social cues and norms. Therefore, the Mindstudio chatbot could act as a temporary means of deterring them from an unethical pathway with quick accessibility.

Support with Addressing Trauma – Pain is universal. If pain is not a part of a living being's day-to-day antics, it is not truly living. People from the targeted group are no exception to this. Studies have shown that the targeted group, especially those with ASD, is more vulnerable to trauma due to their limited self-regulation and sense-making (Buuren, Hoekert, and Sizoo, 2021).

There is even evidence that people in the group in question have much higher exposure to violence, parental divorce, traumatic loss, mental illness, and substance abuse in their families (Buuren, Hoekert, and Sizoo, 2021). Trauma and stress have always found their footing through these means. An increased reign of anxiety over one's mind, especially with the targeted group, can lead to increased introversion and alienation from the surrounding external world. These conditions will eventually result in an increased aversion to interacting with people in regular environments (i.e. school, college, or work). There is no doubt that a psychologist or social worker could do great work here. What if they were to collaborate with the developer to make a Mindstudio application that takes professional input data on how to help people of the targeted group who have dealt with trauma? If a 19-year-old girl named Shalini with ASD experiences psychological abuse from one of her parents due to the use of illegal substances, which then leads to divorce, her ability to think positively and open-mindedly could be substantially hindered. The aforementioned AI system could furnish tips on how to approach the situation rationally and allow gradual healing. The sheer speed of the app with modern APIs can swiftly help her heal with temporary advice before she is able to consult an in-person social worker or psychologist, which of course should be prioritized. It should be remembered that if the information is deemed unsatisfactory by her, depending on the severity of her symptoms, the AI prompt will also provide contact information to professional organizations designed to aid people like her.

Open To Improvement and Innovation – When humans and machines can be made close allies through times of ease and difficulty, this is arguably the most beneficial relationship those parties can have. Improving, updating, and increasing the capabilities of natural language processing as well as reinforcement learning applications is the best way to strengthen that dynamic. There has been evidence of AI applications being used to increase people's knowledge of human behavior, specifically regarding children with autism in educational environments. By utilizing an artificial intelligence system to gain knowledge from the targeted group, various fields that researchers and engineers can explore open multiple new pathways for investigation. If the group in question can have their livelihoods substantially improved at the psychological and social levels using the app, it is possible that an entire effort can be made to improve the lives of the targeted group from an earlier age, even before their teenage or early adult years. This would refer to children with high-functioning ASD or social anxiety as young as six or seven. These ages are usually when children start learning how to hold conversations, rather than just playing games or drawing with their peers. Were this application to be used by reputable organizations such as AccessOAP (Ontario Autism Program), children with high-functioning ASD could use the application to understand the fundamentals of ethical social decisions from an early age. This, in turn, can prevent any sort of turmoil or social conflict as they reach later years. This is a fairly new idea but is it still an intriguing concept for deep-learning applications. There is no doubt that there would need to be a serious change to the aesthetics and presentation of information in the app. With the targeted group you can have relatively simplistic but appealing designs with HTML and CSS, but for younger children, something flashier would be more suitable (Dogra, 2022). Researchers and engineers can take the opportunity to see how they can simplify the style in which it conveys the social etiquette lessons to the kids, such as using vibrant AI-generated drawings with APIs from OpenAI as a foundation. Easily understandable language like the type used in books aimed at children around 6-10 would also be a good tip. There is also the possibility of advancements in AI technology changing how developers create projects similar to the Mindstudio one. One example is VR. If developers could use VR technology to create virtual conversations with AI-powered digital counselors or social workers to provide temporary support in social skills and more for adolescents, it could be very engaging experience. The only drawback though is that VR technology is relatively expensive, limiting their accessibility significantly (Daws, 2024). These are just a few ideas that can be considered as positive and promising outcomes from the discussed AI app for the targeted group.

**Results**

The application was reviewed by Laticia DiBernardo from the organization Reconnect Community Health Services in November of 2024. DiBernardo is a qualified social worker experienced in working with youth on subjects related to mental health struggles and personal crises. DiBernardo gave her opinion on the app's key features below for each one:

Main Menu of the App – DiBernardo felt that the main menu (Figures 11 and 12) was well-structured, clear, concise, and effective in conveying the nature of the app. She felt the introductory text was welcoming but also professional in its style. The clarity of the qualified individual (that being DiBernardo herself) who reviewed the app and her LinkedIn profile link made it easy for the user to feel they can trust the app according to her feedback.

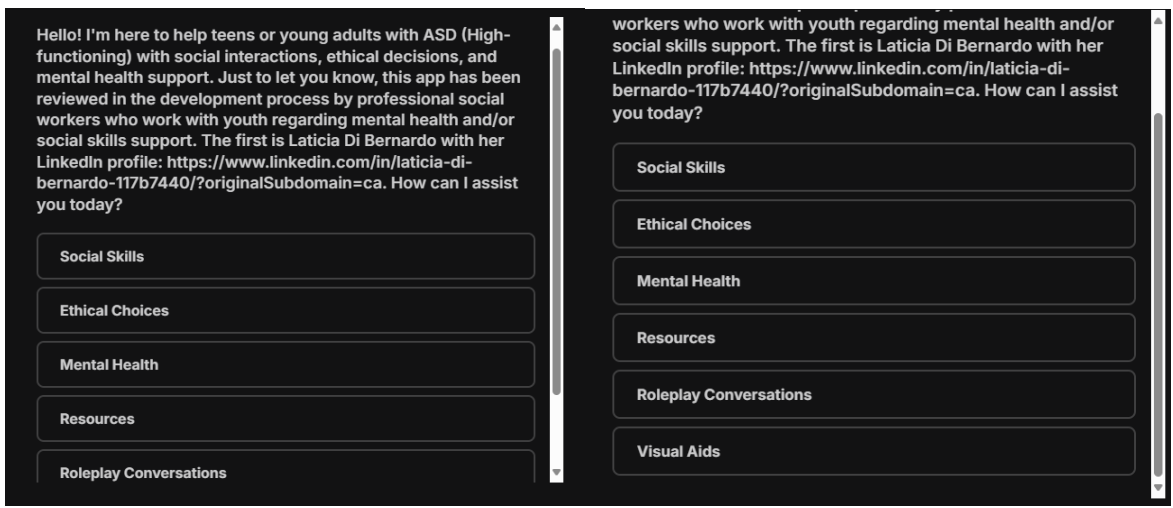

Figures 11 and 12: These show the main page of the application of the app during its professionally reviewed test run.

Social Skills Support – The introductory text was met positively. Mrs. DiBernardo commented that the text is very clear and understandable for users (Figure 13). Then considering the fact that the text does not shy away from advising the user to prioritize in-person support or guidance, it was met positively as well. When the prompt functionality for this section of the app was tested,

DiBernardo found the AI-generated response to be very helpful and detailed. The prompt was about understanding how to speak to new people at school and ways to gain it as a skill gradually. Mrs. DiBernardo found the breakdown of the steps to be the most powerful and effective aspect of the app's capability in this section. She also felt the text-to-speech was very beneficial for engaging users of different learning preferences (Figures 14 and 15).

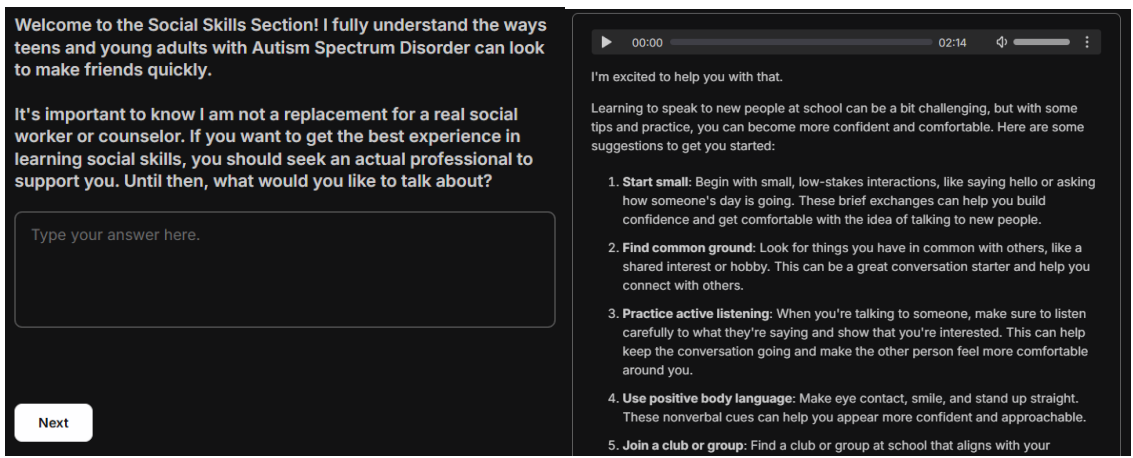

Figures 13 and 14: These show the Social Skills section's introductory text and initial response to the user's prompt.

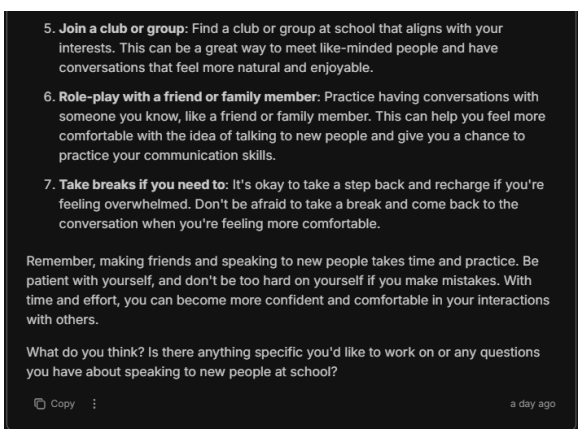

Figure 15: This shows the rest of the initial response to the user's prompt.

Ethical Guidance – This section was met with a very positive response from DiBernardo as well. This regards the opening text's clarity and reinforcement of the importance of seeking real-world guidance over relying solely on the chatbot (Figure 16). The prompt that was made for this section was about a potential user conveying their anger and frustration towards their friends for not being able to hang out more frequently. The app then gives detailed tips on healing and positively finding ways for the user to change their mindset to the benefit of themselves and others (Figures 17 and 18). This again is according to the reaction of Mrs. DiBernardo.

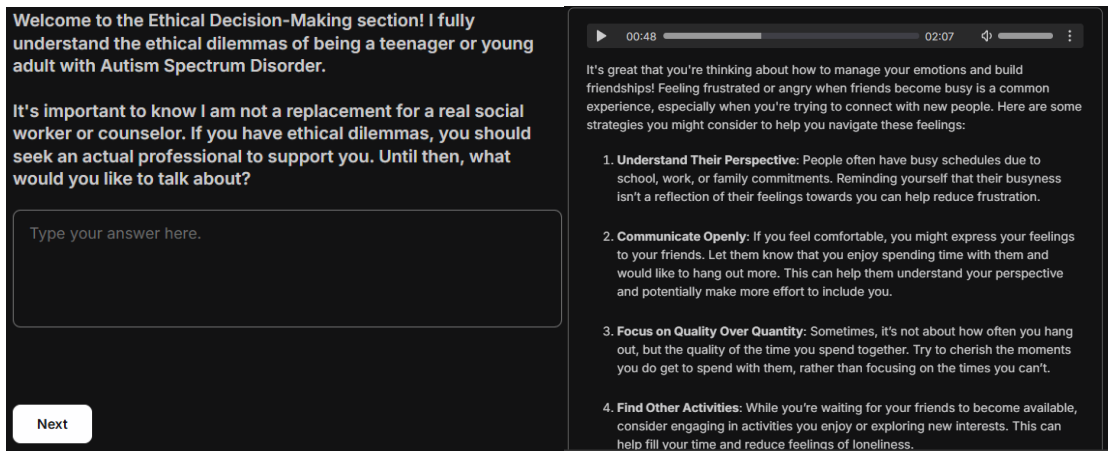

Figures 16 and 17: These show the Ethical Decision section's introductory text and initial response to the user's prompt.

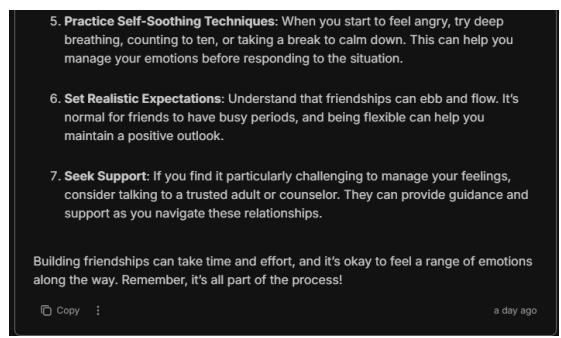

Figure 18: This shows the rest of the initial response to the user's prompt.

Mental Health Support – This part of the chatbot application was positively viewed by DiBernardo. She emphasized the importance of the introductory text here emphasizing seeking real-life counseling specifically for this section since mental health struggles are somewhat different from the other struggles the app is tailored for. Mental health challenges can potentially be life-threatening, therefore having this statement is even more crucial at the start of this section according to Mrs. DiBernardo's professional opinion (Figure 19). The prompt that was used for this test run is about a user feeling sad about their friends from elementary school moving away while the person using the app has to struggle to make new friends in their upcoming high school term. DiBernardo felt the chatbot's response was very uplifting and upbeat in its style (Figure 20). She also thought the caring and constructive nature of the response was effective in helping a potential user have hope that their circumstances would improve over time, therefore making this a helpful temporary tool for its tailored purpose.

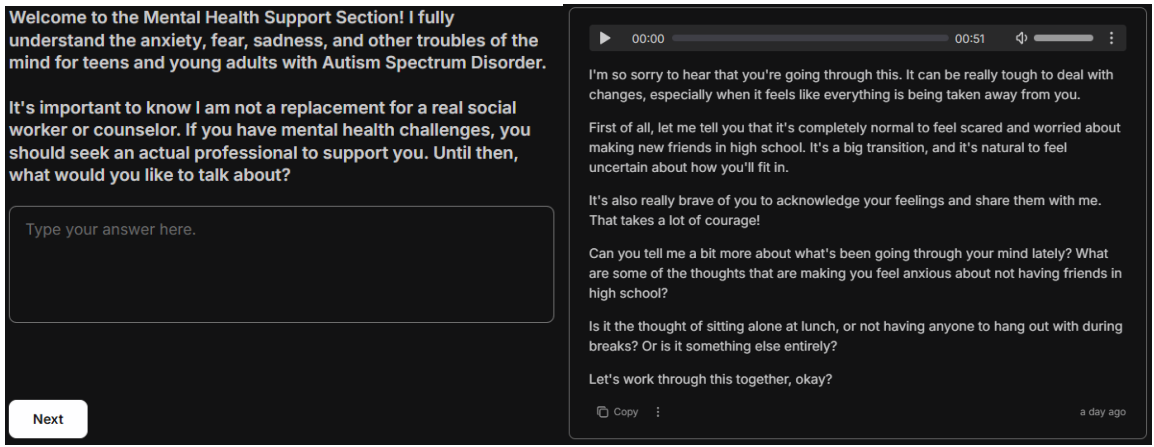

Figures 19 and 20: These show the Mental Health section's introductory text and initial response to the user's prompt.

Resources – Mrs. DiBernardo stated that this section is very useful and convenient for users who want to seek actual support rather than just using the chatbot after seeing the app's textual reminders in the other sections. This test run's prompt was about the application's ability to give the names of different organizations for helping a teenager with ASD and their mother to help her child seek support for social skills development and to make friends. DiBernardo really liked the positive and encouraging words at the end after the list was made by the app (Figures 22 and 23).

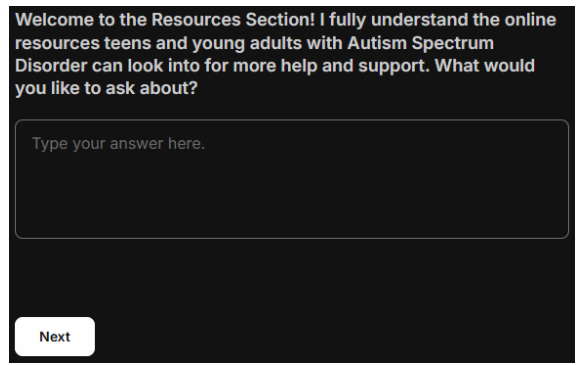

Figure 21: These show the Resources section's introductory text.

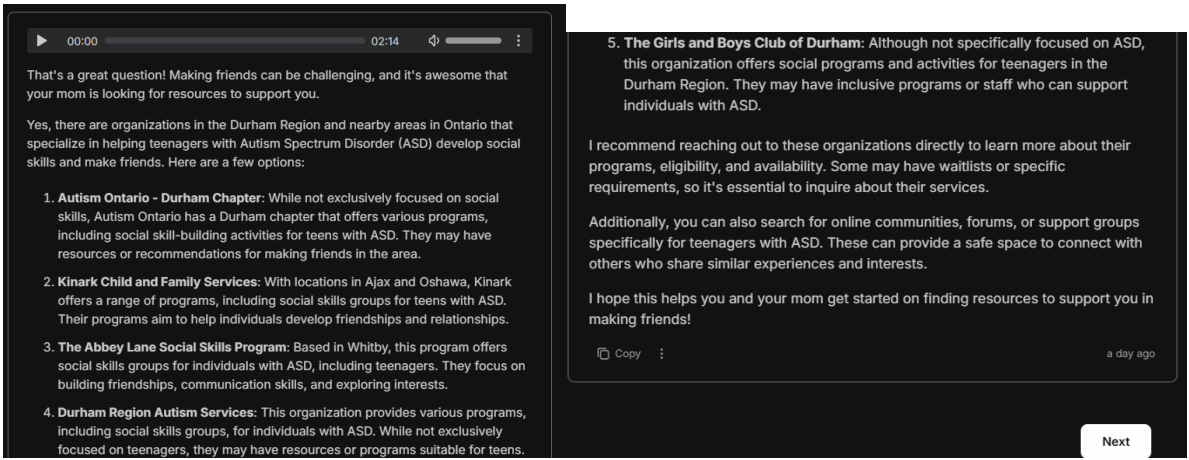

Figures 22 and 23: These show the Ethical Decision section's initial response to the user's prompt.

Roleplay Conversations – This section was met with some positive attention by DiBernardo. She had felt it was a decent side tool for users rather than being the main purpose of the app. This is due to real-life conversations and practicing social skills in that avenue is more crucial than using a chatbot as she stated. She felt the choice to speak to a male or female character was inclusive and good for personalizing it clearly for the user. This is coupled with the branching section for female character roleplay conversations where the user can choose to speak to a classmate, teacher, co-worker, or work boss to create variety. For this test run, we chose the female classmate roleplay (Figure 25). Mrs. DiBernardo liked the easy-going nature of how the app communicates here but also felt it was somewhat unrealistic to how actual conversations occur in a real-world setting (Figure 26). She also felt that this was by no means a large detractor from the quality of the app as the main purpose is to provide a temporary solution to the struggles of many people with ASD.

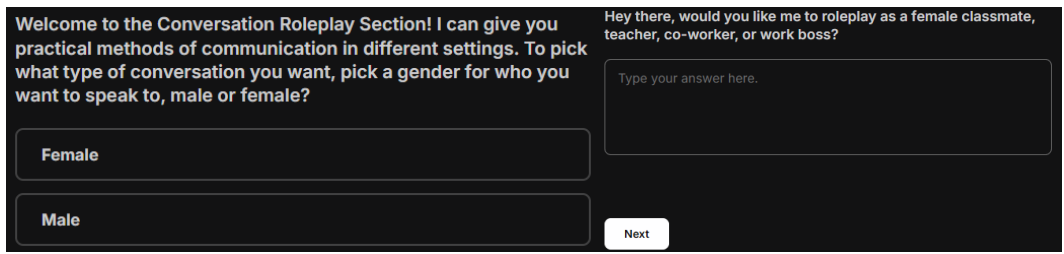

Figures 24 and 25: This shows the Conversation Roleplay section's introductory text and female character creation section upon the user's choice.

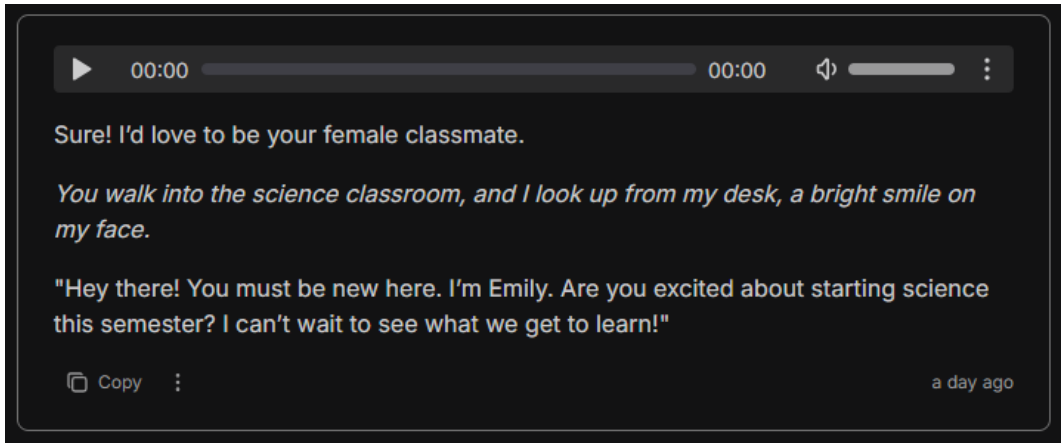

Figure 26: This shows the initial female character's response to the user's prompt.

Visual Aids – This section, which is unique compared to the rest was viewed in a positive light by DiBernardo. The test run included the prompt for this section was about how the user can understand posture that could make them more approachable to potential new friends or colleagues. The generated image in Figure 28 by the application has people of different backgrounds conveying different levels of approachability through their arm posture. Some of them are more reserved and introverted compared to others, Mrs. DiBernardo commented. She appreciated the text underneath the image (Figure 29 and 30) made it easy for the user to comprehend the message of the application in that there are better ways of using posture to help make friends with people than others. The fact that the chatbot specified the importance of open arm posture as more effective in getting people to gravitate towards you was well received by DiBernardo.

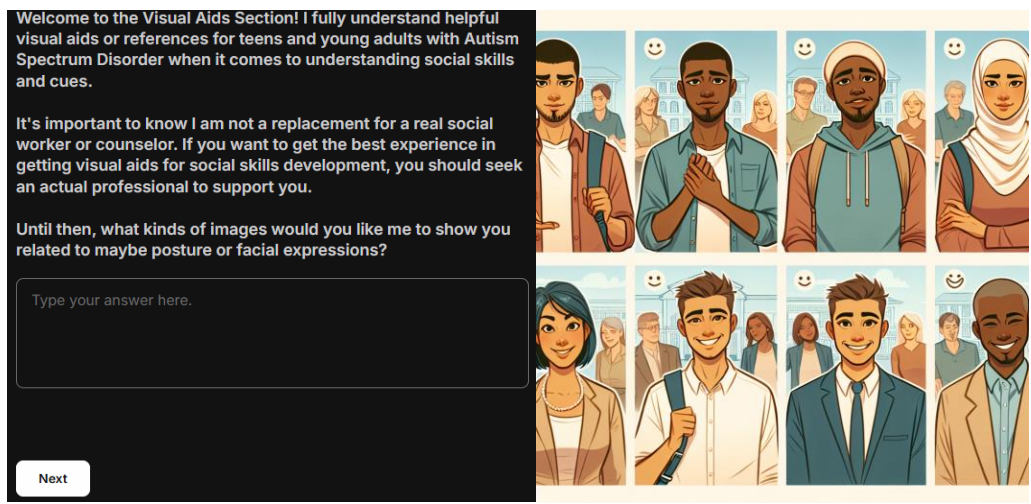

Figure 27 and 28: These show the Visual Aids section's introductory text and initial response to the user's prompt.

![Screenshot of chatbot response about posture and body language tips]

Figures 29 and 30: These show the rest of the initial response to the user's prompt.

Overall, the app was well received by Laticia DiBernardo and she feels that it shows the potential Artificial Intelligence has in guiding and providing support for people of the targeted group. She understood that it is a beneficial tool for people who may not yet have access to in-person counseling with a social worker or psychologist.

**The Limitations**

Table 2: This shows a summary of the limitations of the application –

| Limitation | Explanation |
| --- | --- |
| Potential for Misinterpretation of User Experiences and Needs | AI applications have been known to spread misinformation or incorrect information. The intended app by Mindstudio could possibly fall under the same predicament considering it relies on LLMs from existing AI companies. |
| Privacy and Consent Concerns | As with ChatGPT, people are concerned about what their personal data is being used for when communicating with chatbots. Since the intended application for the targeted people is made in Mindstudio.ai, similar concerns could exist as to whether or not nefarious individuals would hack into the app and face identity theft. |
| Replacement of Real Interactions | Users may prefer the ease of communicating with the chatbot versus taking the challenge of seeking real relationships and friendships. |
| Excessive Reliance Concerns | Users may rely on the app excessively to the point that valuable time is wasted, creativity is diminished, and their mental/physical health is |

|                                  | harmed.                                          |
|----------------------------------|--------------------------------------------------|
| Potential for Exploitation       | If the Mindstudio.ai application were to be hacked by nefarious individuals, they may attempt to mislead users into buying products online as part of a scam operation. |
| Clean Text in Image Generation   | The application runs into similar issues as ChatGPT with creating clear and coherent text in AI-generated images. |

Potential for Misinterpretation of User Experiences and Needs – Regarding the creation of an application to present socially ethical advice and mental health support to the targeted group, there are many potential drawbacks. One is the AI chatbot providing misleading information to teens or young adults using it. These individuals are hearing many of these tips for the first time, especially if they are inexperienced in conversing or socializing in their own academic and workplace environments. AI-powered chatbots have been documented to spread misinformation when the tools used to create them are not regulated and are available to the public without much oversight (Virginia Tech News, 2024). This is the equivalent of a new journal forum that is completely open to anonymous input that can be inserted with impunity. Falsehood comes from these circumstances as orange does from red and yellow. There have been situations in which AI has been allowed to generate false information via Language Learning Models (LLMs) for fake news sites (Virginia Tech News, 2024). While this is understandably alarming, we should also consider the ways developers and engineers work to counter this predicament. AI detection tools are being implemented and regularly utilized by companies such as major news corporations (Endert, 2024). The best way for the Mindstudio AI chatbot to counter false information is, as stated before, to ensure the application has the most correct information it can with the oversight of a professional psychologist or social worker. One of the reasons this is crucial is that if distorted or pernicious information is conveyed to the targeted group regarding their psychological woes, it could open the gate to exacerbated stress, sorrow, and more. As an example, if a 20-year-old man with High-Functioning ASD named Fernando felt inundated with grief over the turmoil of a healthy and lasting friendship, he would understandably seek a way out of it. If he had a social worker who recommended the proposed AI chatbot whenever they were not available for an appointment with Fernando, he would have used the app. If the app gives him tips that could confuse him with contradictory information due to a lack of adequate testing or reviewing, Fernando would be distressed. This, in turn, can bring about a lack of trust in the organization that recommends the app in the first place.

Privacy and Consent Concerns – All people, whether they are part of the targeted group or not, should fundamentally have the right to guard their personal information and give consent to potentially life-changing transactions or interactions. This is arguably conducive to maximizing societal well-being. The data used to fuel AI chatbots comes from the general public, from what we say and do, and how they are presented online. With the AI chatbot designed to provide mental health support and tips for social ineptitude to the group in question, there are some key factors to be perturbed. This AI system would be designed to collect the deepest personal thoughts, behavioral displays, and emotional tangents of the targeted group. The developer must take the extra steps needed to prevent any sort of data leakage or unknown agents from accessing it. Since the application itself is being made by Mindstudio.ai, it seems to fall under a similar level of concern as other popular chatbots such as ChatGPT. This is due to Mindstudio utilizing LLM models for usage such as GPT. A partner from BDB Pitmans (a high-profile law firm in the UK) has stated that while personal data is not used to profile users of ChatGPT, people understandably find it intrusive (O'Flaherty, 2024). The same person expressed that while OpenAI restricts user data to train their models, it is less clear what OpenAI does for people objecting to it disclosing their private information in chat responses (O'Flaherty, 2024). The best way to traverse these concerns would be a precise effort to only grant access to the application to people of the targeted group who have explicitly been diagnosed with ASD or general social anxiety by a professional psychologist. If this cannot be done, then there is the safety of usernames and passwords. A robust account authentication system with these two would be ideal to prevent anonymous individuals from accessing people's accounts. This is what Mindstudio integrates for developers and users, in addition to allowing for other options. One of these is using one's own Gmail account authentication information to log in to Mindstudio. When considering Gmail users can add multiple authentication layers to protect their personal information (credit cards, SIN, etc.), privacy becomes fairly promising. Private information being leaked or accessed by unauthorized agents would cause great stress and anxiety for the individual in the group in question. The trust between people and the app would be tarnished resulting in more apprehension for them to seek guidance or support from external forces. People using the app would have to be made to clearly understand what they are consenting and agreeing to before they begin chatting with the app, especially if they are minors. Parental consent would also be needed, especially if organizations focused on working with the group in question were to regularly recommend it to clients.

Replacement of Real Interactions – Human connection has been scientifically proven to be a cornerstone of the well-being of all of us psychologically (Soler-Gonzalez, San-Martin, Delago-Bolton, and Vivanco, 2017). The need for social connection is like our need for food when we're hungry. A need like this is likely a modern evolutionary product. Neuroscientist Matthew Lieberman is mentioned to have said our minds have evolved through the ages to naturally anticipate interacting with others and forming relationships (The Social Creatures, 2024). Were people to suddenly abandon this, no doubt it would cause great confusion and disorientation, not to even mention the certain seeds of sad turmoil it would plant. What else could be expected from rejecting our intuitive desires? Mass agitation of people's minds has never been advantageous to a healthy, long-lasting, multi-generational community. This is arguably one of the multiple dangers of giving an AI chatbot for the purposes previously described to the group in question. The targeted group clearly struggles with social confidence and having a sense of ease in very public landscapes. Would there not be a great risk of them isolating themselves from their peers at high school and college from the sheer comfort of speaking to a realistically characterized algorithm? The developer of the Mindstudio project would have to investigate designing the app to have limits. These limitations would entail the limits as to how realistic the personality of the app can be. If this cannot be done, it presents reasonable concerns. Machine Learning and Deep Learning applications have been noted as becoming increasingly sophisticated with tasks with significant societal effects (Fan, Yan, and Wen, 2023). If the application became so lifelike in inhabiting a real psychologist's or social worker's personality, the person (from the targeted group) using it would possibly not want to stop and (perhaps unwittingly) assign friend-like attachments to it, especially since the AI system would be designed to create ease of the mind. The chatbot itself does not stop functioning in a fully realistic manner after giving the needed social advice or mental health tips to the individual. It could be programmed to then encourage the reader to apply the teachings to real-life scenarios or at least provide a reminder at the start of conversations to seek real relations and professional help. The app can also have limits as to how many interactions the user can have with the app before they need more tokens given to them by the developer. This way not only would the targeted people get the easily accessible advice they may need, but an external force would bolster them to get the courage to apply the knowledge they've gained.

Excessive Reliance Concerns – This naturally ties to the point above as a lack of proper, healthy human connection and personal harm to individuals can be conjured by compulsive usage of AI apps. OpenAI's chief technology officer Mira Murati has recognized this (Klara, 2023). Compulsive interactions with artificial intelligence can open the door to multiple societal burdens. Some examples are waning physical/mental health, time-wasting, and the cessation of creativity (Cunningham, The Joint, n.d.). AI chatbots require you to interact with a screen obviously, so similar symptoms from screen addiction can emerge here as well. Headaches, sleep deprivation, and weight gain could all occur. Not to mention the effect of excessive solitude on one's psyche (Cunningham, The Joint, n.d.). Along with this, critical thinking could gradually diminish through repeated use of AI for even solving simple math problems or other tasks that should be manually investigated with personal creativity. Indeed, these technological devices can help with many tasks that regularly require deep logical assessment, but what if we find ourselves in situations without access to AI systems such as exams, business pitches, or job interviews? Our weakened creativity and ability to think on our feet would come back to haunt us. If we take these concerns and see the effects on the group in question, there are many reasons to doubt the beneficial side of the previously proposed AI chatbot. Keeping in theme with the previous point that the targeted group can easily become attached to the previously discussed AI chatbot out of fear of in-person interactions or conversations, they can suffer the above-mentioned harms as well. One of the easiest ways to deal with this threat is to instill limited usage with the AI chatbot. Mindstudio.ai allows developers to limit usage via tokens for each user. Therefore, the developer of this specific app could substantially lower the app's usage to prevent all-day or even multiple hours of usage each day. When someone from the targeted group goes to speak with a psychologist or social worker, they have timed appointments. The opening pages of each option on the app's menu (social skills support, mental health support, ethical advice) can convey professionally approved words of uplifting motivation for the user. There can also be text explaining that this app is not a replacement for real psychologists or social workers, rather it is a temporary tool for support before the user can seek in-person help. This way, the person from the targeted group using it will have the reminder of the importance of not only seeking a real psychologist or social worker at the first chance but also for going with wisdom and intelligence into their environment (i.e. school, university, workplace) to seek healthy bonds with those around them.

Potential for Exploitation – AI has put many people's livelihoods on the line. It can be used for nefarious purposes and to manipulate the minds of its users. Bots powered by AI can make this harder to sense and counter.   This is especially powerful on social media. The group in question is also vulnerable to exploitation's grip. The chatbot application that has been created and discussed earlier would intake personal data from its users. If the data were to be implicitly taken and handed around anonymously, it would open the floodgates to the targeted group being taken advantage of. If some not-so-honorable individuals were to hack into Mindstudio and gain control of this app, they may try to manipulate the emotional leanings of the targeted group using orchestrated tactics. These could include pushing specific services or products from collaborating companies or even businesses belonging to the developers. These developers may attempt this with in-app purchases, and extra subscription models that bait the group in question with promised "special features" for enhanced support or advice but in reality, make little to no difference other than decreased bank savings as well as trust for the user. People from the targeted group are particularly in danger in a scenario like that since they naturally are less knowledgeable about what is considered socially or ethically acceptable, so they can easily be manipulated into doing things like this at their expense while filling in the pockets of the hackers and businesses involved in the product/service promotions. Their anxiety may decrease their chances of seeking confirmation from their family as to the nature of what they are partaking in. The only foreseeable way this could all occur is if Mindstudio.ai was to be hacked, but up until now, there are no records of such an event. At the same time, there is no guarantee that it will never happen. The application itself will be reviewed for signs of unethical design by a psychologist or social worker during its production. Having extensive monitoring from parental/guardian figures is highly beneficial for the young adults using the app. The parent or guardian should monitor the activity as well as investigate the app for any signs of financial exploitation in the possible scenario that the application has been hacked. Without an installed and procured safety net, the group in question could be vulnerable to mistreatment.

Clean Text in Image Generation – Since the Mindstudio app uses Dalle from OpenAI to generate visual aids, it carries on the same deficiencies that Dalle presents through ChatGPT. The most notable one is the lack of clean readable text within images. This could be an issue if someone using the app would like a helpful image with labels or titles in it for more engagement as the text itself is usually incomprehensible. It is important to consider though that the chatbot includes descriptive text underneath the image generated via the LLM model used (e.g. Llama, Mistral, GPT).

**Conclusion**

The final outlook of this application in terms of its benefits to the targeted people as well as the clear limitations it presents, is that the application's purpose was reasonably fulfilled. "A-Way-Out.ai" at its core was meant to be a research and proof-of-concept undertaking with professional oversight/input, as well as demonstrated effectiveness. The app's construction methodology ensured easy development, testing, and a user-friendly interface. It has shown that Artificial

Intelligence can be used for the social and psychological betterment of people in the targeted group.

**Acknowledgments**

I would like to give special thanks to Professor Qusay Mahmoud from Ontario Tech University and Laticia DiBernardo from Reconnect Community Health Services for their guidance as well as assistance.